\documentclass[twoside]{article}

 \usepackage[accepted]{aistats2021_modified}
\usepackage{algorithmicx}
\usepackage{algpseudocode}
\usepackage{booktabs}
\usepackage{amsmath,amssymb,amsthm, amsfonts}
\usepackage{stmaryrd}
\usepackage{bbm}
\usepackage{float}
\usepackage{pifont}
\usepackage{stmaryrd}
\usepackage{graphicx}
\usepackage{xcolor}
\usepackage{algorithmicx}
\usepackage{algpseudocode}
\usepackage{amsmath,amssymb,amsthm, amsfonts}
\usepackage{stmaryrd}
\usepackage{bbm}
\usepackage{float}
\usepackage{stmaryrd}
\usepackage{graphicx}
\usepackage{xcolor}
\usepackage[round]{natbib}

\definecolor{linkblue}{rgb}{0.10,0.40,0.70}

\usepackage[colorlinks=true,citecolor=linkblue]{hyperref}

\newcommand{\OT}{\mathrm{OT}}

\newcommand{\vect}{\mathrm{vec}}

\newcommand{\E}{\mathbb{E}}

\newcommand{\Ham}{\mathcal{H}}

\newcommand{\C}{\mathrm{\Pi}}

\newcommand{\V}{\mathbb{V}}
\newcommand{\R}{\mathbb{R}}
\newcommand{\pr}{\mathrm{Pr}}

\newcommand{\cmark}{ \textcolor{green!60!black}{\ding{51}}}
\newcommand{\xmark}{ \textcolor{red!60!black}{\ding{55}}}

\DeclareMathOperator*{\argmin}{arg\,min}
\DeclareMathOperator*{\argmax}{arg\,max}

\begin{document}

%

%

\twocolumn[

\aistatstitle{Bayesian Inference for Optimal Transport with Stochastic Cost}

\aistatsauthor{ Anton Mallasto \And Markus Heinonen \And  Samuel Kaski }

\aistatsaddress{Department of Computer Science, \\Aalto University, Finland\\ \texttt{anton.mallasto@aalto.fi} \And Department of Computer Science, \\ Aalto University, Finland \\ \texttt{markus.o.heinonen@aalto.fi}\And Department of Computer Science, \\ Aalto University, Finland \\ Department of Computer Science,\\ University of Manchester, UK \\ \texttt{samuel.kaski@aalto.fi}} ]

\begin{abstract}
  In machine learning and computer vision, optimal transport has had significant success in learning generative models and defining metric distances between structured and stochastic data objects, that can be cast as probability measures. The key element of optimal transport is the so called lifting of an \emph{exact} cost (distance) function, defined on the sample space, to a cost (distance) between probability measures over the sample space. However, in many real life applications the cost is \emph{stochastic}: e.g., the unpredictable traffic flow affects the cost of transportation between a factory and an outlet. To take this stochasticity into account, we introduce a Bayesian framework for inferring the optimal transport plan distribution induced by the stochastic cost, allowing for a principled way to include prior information and to model the induced stochasticity on the transport plans. Additionally, we tailor an HMC method to sample from the resulting  transport plan posterior distribution.
\end{abstract}

\section{INTRODUCTION}\label{sec:intro}
Optimal transport (OT) is an increasingly popular tool in machine learning and computer vision, where it is used to define similarities between probability distributions:  given a \emph{cost function} between samples (e.g. the Euclidean distance), representing the cost of transporting one sample to another, OT extends it to a cost of transporting an entire distribution to another. This \emph{lifting} of the cost function to the space of probability measures is carried out by finding the \emph{OT plan}, which carries out the transport with minimal total cost.

\begin{figure*}
\centering
\includegraphics[width=.8\textwidth]{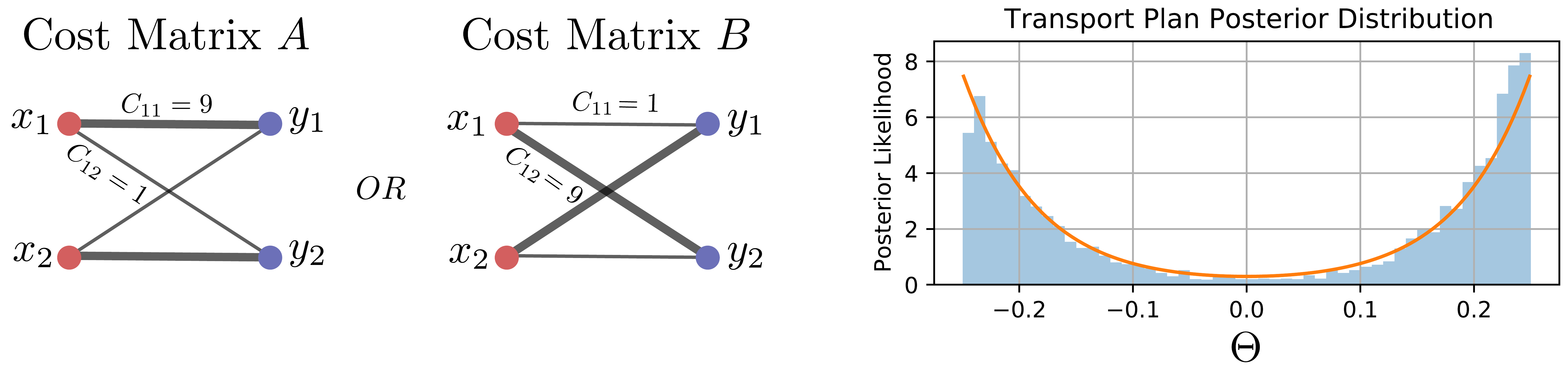}
\caption{OT with stochastic cost. Assume measures $\mu,\nu$ having uniform distribution over the atoms $x_1,x_2$ and $y_1,y_2$, respectively, and either of the cost matrices $A$ or $B$ is observed with equal probability, so that on average $C_{ij}=5$ for all $i,j$. The average cost matrix yields an ill-posed OT problem, as any transport plan would solve the OT problem. On the other hand, the posterior distribution for the transport plan (on the right, blue gives an empirical histogram for posterior samples, orange gives posterior likelihood) encaptures the multimodality, which arises, as there are only two minimizing transport plans for the problem, depending on whether we witness cost matrix $A$ or $B$. The transport plan is a $2\times 2$ matrix, but can be parameterized with a single real value $\Theta$.}
\label{fig:intro}
\end{figure*}

Traditional OT assumes a deterministic and exact cost between samples~\citep{villani08,peyre19}. This is natural for most of OT applications in machine learning, such as defining loss functions for learning probability distributions, e.g., in Wasserstein generative adversial networks (WGANs)~\citep{arjovsky17}, or defining statistics for stochastic data objects, e.g., between Gaussian processes representing random curves~\citep{mallasto17}. However, the assumption of an exact cost rarely holds in real-life OT applications, such as logistics on real-life road networks, or when the transported distributions vary spatially. See Fig.~\ref{fig:intro} for an illustration.

As the transportation cost varies, a natural question arises: how to take this uncertainty into account in the transportation plan, and which of them should be used in practice? Furthermore, it is important to include any prior knowledge in the solution. To answer these questions, we propose to use the Bayesian paradigm in order to infer the distribution of transport plans induced by the stochastic cost, and name the resulting approach as BayesOT. As a special case, we show that resulting point estimates for the OT plan correspond to well-known regularizations of OT.

We contribute
\begin{itemize}
    \item[1.] BayesOT, A Bayesian formulation of the OT problem, which produces full posterior distributions for the OT plans, and allows interpreting known regularization approaches of OT as maximum a posteriori estimates.
    \item[2.] An approach to solving OT problems having stochastic cost functions between samples of the two marginal measures.
    \item[3.] A Hamiltonian Monte Carlo approach for sampling from the transport polytope, i.e., the set of joint distributions with two fixed marginals.
\end{itemize}

\textbf{Related Work.} We are not aware of earlier works on stochastic costs in OT, but some works are related. For example, \emph{Schrödinger bridges} consider the most likely path of evolution for a gas cloud, given an initial state and an evolved state, a problem equivalent to entropy-relaxed OT~\citep{dimarino19}. The evolution is Brownian, thus the dynamics bring forth a stochastic cost; however, no stochasticity remains after the most likely evolution is considered. 

\emph{Ecological inference}~\citep{king04} studies individual behavior through aggregate data, by inferring a joint table from two marginal distributions: this is precisely what is done in OT, using the cost function. \cite{frogner19} consider a prior distribution over the joint tables, and then compute the maximum likelihood point estimate. Our work is related, as our approach, in addition to the prior distribution, adds a likelihood, relating the joint table to the OT cost matrix. These two components then allow computing \emph{maximum a posteriori} (MAP) estimates and to sample from the posterior in a Bayesian fashion. \cite{rosen01} consider Markov Chain Monte Carlo (MCMC) sampling from a user-defined prior distribution to estimate the joint table. However, strict marginal constraints are not enforced, which \cite{frogner19} speculate is due to the difficulty of MCMC inference on the set of joint distributions with perfectly-observed marginals. In contrast, BayesOT takes the marginal constraints strictly into account.

A conceivable alternative approach to solving the OT problem with stochastic cost would be to use standard OT,
applied on the average cost. An obvious down-side of this approach would be losing all stochasticity, resulting in an average-case analysis. If the measures are hierarchical, i.e., we have mass distributions $\mu_i,\nu_j$ over spatially varying components given by random variables $X_i, Y_j$. Then, the cost $c(X_i, Y_j)$ would be stochastic, depending on the realisations of the components. One could then consider extending the sample-wise cost to a component-wise cost using the OT quantity between the two components, i.e., $\tilde{c}(X_i, Y_j) = \OT_c(X_i, Y_j)$~\citep{chen18}. However, we would lose all stochasticity again, and the component-wise OT cost would be blind to any natural correlation between the components.

Furthermore, one could solve the OT plan associated with each cost matrix sample $C^k$, and carry out population analysis. This would, however, prevent the use of prior information, and no likelihood information would be given on the OT plans, which could be used to estimate the relevancy of a given plan. 

\begin{table}[]
    \centering
    \resizebox{\columnwidth}{!}{%
    \begin{tabular}{lcccc}
        \toprule
          &  \multicolumn{2}{c}{Cost} & & \\
          \cmidrule(lr){2-3}
          & exact & stochastic & Prior & Uncertainty  \\
         \midrule
         OT &\cmark &\xmark &\xmark & \xmark\\
         RegularizedOT & \cmark & \xmark & \cmark & \xmark\\
         BayesOT & \cmark & \cmark & \cmark & \cmark\\
         \bottomrule
    \end{tabular}
    }
    \caption{Comparison between vanilla OT, Regularized OT, and our method BayesOT. The qualities imply whether the approaches are able to incorporate an exact or stochastic cost, prior information, or whether the methods provides uncertainty estimates.}
    \label{tab:my_label}
\end{table}

\section{BACKGROUND}\label{sec:background}
We now summarize the basics of OT and Bayesian inference in order to fix notation.

\textbf{Optimal Transport} is motivated by a simple problem. Assume we have locations of factories $\{x_i\}_{i=1}^n\subset \R^d$ and of outlets $\{y_j\}_{j=1}^m\subset \R^d$ in the same space. Each of the factories produces $\mu_i$ amount of goods, and the outlets have a demand of $\nu_j$, each positive and normalized to sum to one; $\mu_i,\nu_j\geq 0$ and $\sum_{i=1}^n\mu_i = \sum_{j=1}^m \nu_j = 1$. We represent the distribution of goods over the factories and demands over the outlets by the discrete probability measures
\begin{equation}
\mu(x) = \sum_{i=1}^n\mu_i\delta_{x_i}(x),~\nu(y)=\sum_{j=1}^m \nu_j\delta_{y_j}(y),
\end{equation}
where $\delta_x(y)$ stands for the Dirac delta function. 

Assume that the cost of transporting a unit amount of goods from $x_i$ to $y_j$ is $c(x_i,y_j)$, where $c:\R^d\times \R^d \to \R^+$ is the \emph{cost function}. Then, the \emph{optimal transport quantity} between $\mu$ and $\nu$ is given by
\begin{equation}\label{eq:OT}
\begin{aligned}
    \OT(\mu, \nu, C) &= \min\limits_{\Gamma \in \C(\mu, \nu)} \OT( C, \Gamma) \\
    &\overset{\Delta}{=} \min\limits_{\Gamma \in \C(\mu, \nu)} \sum_{i,j} \Gamma_{ij}C_{ij},
\end{aligned}
\end{equation}
where we have the set of joint probability measures with marginals $\mu$ and $\nu$,
\begin{equation}
    \C(\mu,\nu)
    \overset{\Delta}{=}\left\lbrace 
    \Gamma: \sum_{j=1}^m \Gamma_{ij} = \mu_i,~ \sum_{i=1}^n \Gamma_{ij} = \nu_j,
    \right\rbrace
\end{equation}
also known as the \emph{transport polytope}. Its elements are \emph{transport plans}, as $\Gamma_{ij}$ is the amount of mass transported from $x_i$ to $y_j$. The \emph{cost matrix} is given by $C_{ij}=c(x_i,y_j)$. The constraints on $\C(\mu,\nu)$ enforce the preservation of mass in the transportation problem; all the goods from the factories need to be transported so that the demand of each outlet is satisfied.

This seemingly practical problem produces a geometrical framework for probability measures, by lifting the sample-wise cost function $c$ to a similarity measure $\OT(\mu,\nu,C)$ between the probability measures. Depending on the cost function, a metric distance could be produced (i.e., the $p$-Wasserstein distances), which allows studying probabilities using metric geometry. Refer to \cite{villani08} for more details on OT, and \cite{peyre19} for computational aspects.

\textbf{Regularized Optimal Transport.} The OT problem in~\eqref{eq:OT} is a convex linear program, often producing slow-to-compute, 'sparse' transport plans that might not be unique. This has motivated regularized versions of OT, which admit unique solutions. We now summarize regularized OT, as it turns out that solving certain \emph{maximum a posteriori} estimates under the BayesOT framework is equivalent to solving regularized OT, as will be discussed in Sec.~\ref{sec:map_estimate}. 

Given a strictly convex regularizer $R$, the $\epsilon$-regularized OT problem is given by~\citep{dessein18}
\begin{equation}\label{eq:rot_mover}
\OT_R(\mu,\nu, C) = \min\limits_{\Gamma \in \C(\mu, \nu)} \lbrace \OT( C, \Gamma)
+ \epsilon R(\Gamma)
\rbrace,
\end{equation}
where $\epsilon > 0$. With some technical assumptions on $R$, such as strict convexity over its domain, there exists a unique minimizer $\Gamma^*$ of \eqref{eq:rot_mover}, which in practice can be solved using \emph{iterative Bregman projections}.

A popular choice for the regularizer is given by $R=H$~\citep{cuturi13}, where
\begin{equation}
H(\Gamma)= - \sum_{ij} \Gamma_{ij} \log \Gamma_{ij},
\end{equation}
is the \emph{entropy}. This specific regularization strategy has gained much attention, as it is fast to solve with the \emph{Sinkhorn-Knopp iterations}~\citep{knight08}, and enjoys better statistical properties compared to vanilla OT~\citep{genevay19}.

\textbf{Transport Polytope.} To accommodate the somewhat complicated constraints on the transport polytope, we cast the polytope as a set concentrated on an affine plane bound by positivity constraints. This allows parameterizing the polytope using a linear chart, as introduced below in \eqref{eq:Gamma_chart}, which will later on be utilized in sampling viable transport plans.

\begin{figure}[b]
    \centering
    \includegraphics[width=.6\linewidth]{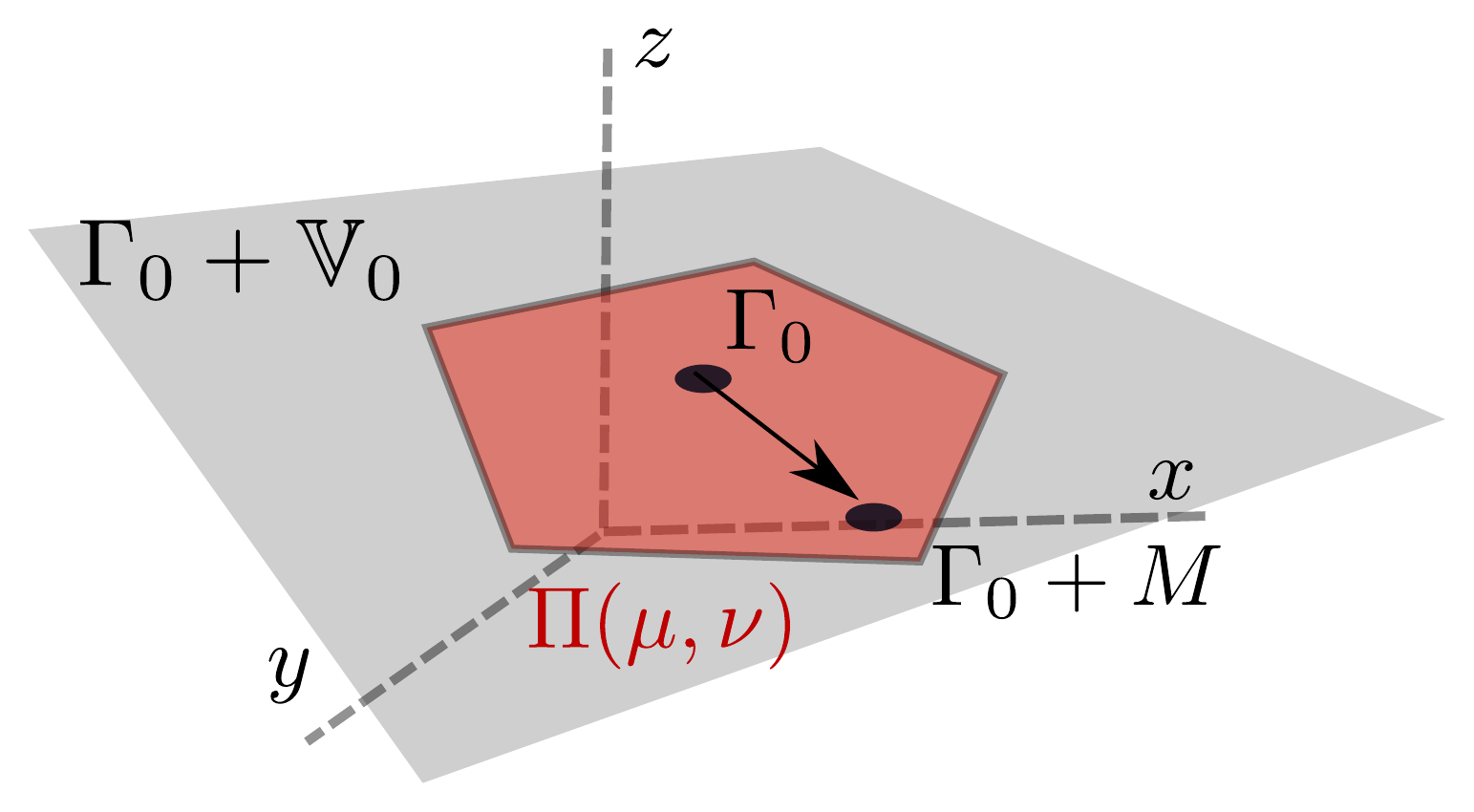}
    \caption{Illustration of the transport polytope as described in \eqref{eq:tpolytope_as_affine}.
    \label{fig:OT_plan_space}}
\end{figure}

Rigorously, we can formulate the constraints on $\C(\mu,\nu)$ in a linear fashion as
\begin{equation}\label{eq:Gamma_constraints}
\begin{bmatrix}
    \Gamma & 0 \\ 0 &\Gamma^T
    \end{bmatrix}\begin{bmatrix}
    \mathbbm{1}_m\\ \mathbbm{1}_n
    \end{bmatrix}
    = \begin{bmatrix}
    \mu \\ \nu
    \end{bmatrix}, \Gamma_{ij} \geq 0,~\forall i,j.
\end{equation}
where $\mathbbm{1}_n$ is the $n$-vector with all coordinates $1$. Hence,  $\C(\mu,\nu)$ is a convex polytope, and furthermore, it lies on the affine plane (see Fig.~\ref{fig:OT_plan_space})
\begin{equation}\label{eq:tpolytope_as_affine}
    \Gamma_0 + \V_0 = \left\lbrace
    \Gamma_0 + M:\sum_{j} M_{ij} = 0,~\sum_{i} M_{ij} = 0,~\forall i,j
    \right\rbrace,
\end{equation}
for some $\Gamma_0 \in \C(\mu,\nu)$. Thus, given any $\Gamma\in \C(\mu,\nu)$, we can find $M\in \V_0$, so that $\Gamma = \Gamma_0 + M$. The vector space $\V_0$ is isomorphic to $\R^{(n-1)\times(m-1)}$ via
\begin{equation}
\begin{aligned}
    \varphi :& \R^{(n-1)\times (m-1)} \to \V_0,\\
    &\Theta \mapsto \begin{bmatrix}
    \Theta & \Theta^R\\
    (\Theta^C)^T & \sum_{ij} \Theta_{ij}
    \end{bmatrix},
\end{aligned}
\end{equation}
where $\Theta^R_i = \sum_j \Theta_{ij}$ is the row sum vector of $\Theta$ and $\Theta^C_j= \sum_i \Theta_{ij}$ is the respective column sum vector. Thus, $\varphi$ provides a linear chart for $\C(\mu,\nu)$ through
\begin{equation}\label{eq:Gamma_chart}
    \Gamma(\Theta) = \Gamma_0 + \varphi(\Theta)\geq 0,
\end{equation}
where the inequality is enforced for all coordinates.

In practice, we choose $\Gamma_0$ to be the independent joint distribution of $\mu$ and $\nu$.

\textbf{Bayesian Inference.}

Assume we are given a family of models $f_\theta$, with parameter $\theta$, and a dataset $D=\{(x_i,y_i)\}_{i=1}^n\subset X\times Y$, produced by an underlying relationship
\begin{equation}
y_i=f(x_i) + \varepsilon_i,
\end{equation}
where $\varepsilon_i$ is a random noise variable, and we want to infer $f:X\to Y$. Given some knowledge about $\theta$ in the form of a \emph{prior distribution} $\theta \sim \pr(\theta)$, Bayesian statistics approaches inferring $f$ by conditioning the parameters via the \emph{Bayes' formula}
\begin{equation}
\pr(\theta| D) = \frac{\pr(D|\theta)\pr(\theta)}{\pr(D)},
\end{equation}
where $\pr(\theta| D)$ is the \emph{posterior distribution}, $\pr(D)$ is the evidence, which can be viewed as a normalizing constant for the posterior distribution, and $\pr(D|\theta)$ is the \emph{likelihood}, given by
\begin{equation}
\pr(D|\theta) = \prod_{i=1}^n\pr(\varepsilon_i = y_i - f_\theta(x_i)).
\end{equation}
The posterior distribution can then be used to estimate the uncertainty of predictions $y = f_\theta(x)$ by sampling $\theta \sim \pr(\theta|D)$ and observing the induced distribution of $y$. This distribution can also be summarized as a \emph{point estimate}. A common point estimate for $f(x)$ is given by the \emph{maximum a posteriori} (MAP) estimate $f_{\theta^*}(x)$, where $\theta^* = \argmax\limits_{\theta} \pr(\theta | D)$. Another popular point estimate is given by the average prediction $\E_{\theta\sim \pr(\theta \mid D)} f_\theta(x)$.

\section{BAYESIAN INFERENCE FOR OPTIMAL TRANSPORT}\label{sec:hier_ot}

We now detail our approach, BayesOT, to solving OT with stochastic cost via Bayesian inference. First,  we motivate the stochastic cost in Sec.~\ref{sec:ot_with_stoch_cost}, and then formulate the problem from a Bayesian perspective in Sec.~\ref{sec:ot_bayes_formulation}. We then focus on sampling from the resulting posterior distribution of OT plans in Sec.~\ref{sec:posterior_sampling}, by devising a Hamiltonian Monte Carlo approach. Finally, we discuss resulting \emph{maximum a posteriori} estimates and their connections to regularized OT in Sec.~\ref{sec:map_estimate}.

\subsection{Optimal Transport with Stochastic Cost}\label{sec:ot_with_stoch_cost}
Consider the scenario where instead of an exact cost matrix, we observe samples $C^k\sim \pr(C)$, $k=1,...,N$, from a stochastic cost $C$, which we view as a random variable. This stochasticity propagates to the OT plan $\Gamma$ via the OT problem
\begin{equation}\label{eq:stochastic_gamma}
\Gamma \sim \argmin\limits_{\Gamma \in \C(\mu,\nu)} \OT(C,\Gamma),\quad C\sim \pr(C).
\end{equation}
In the rest of this work, our goal is to infer the distribution $\Gamma$ inherits from $C$.

Stochastic costs naturally occur when considering OT between hierarchical models $(\mu_i,X_i)_{i=1}^n$ and $(\nu_j,Y_j)_{j=1}^m$, where $X_i$, $Y_j$ are random variables taking values in $\R^d$, resulting in the stochastic cost matrix $C_{ij}\sim c(X_i, Y_j)$. Here one can understand $(\mu_i, X_i)$ as a mobile factory with mass $\mu_i$, that has a stochastic location according to the random variable $X_i$.

On the other hand, the cost $c$ can inherently be stochastic, e.g., when transporting goods in real life, as traffic congestions behave stochastically, affecting the cost of transporting mass from point $i$ to point $j$.

The Bayesian choice to tackle~\eqref{eq:stochastic_gamma} provides a convenient way of expressing uncertainty in parameters, allows the inclusion of prior knowledge on $\Gamma$, alleviating problems with sample complexity, and provides a principled way of choosing point-estimates as the \emph{maximum a posteriori} (MAP) estimates.

\subsection{Bayesian Formulation of OT}\label{sec:ot_bayes_formulation}
To employ Bayesian machinery, we need to define a prior distribution $\pr(\Gamma\mid \mu,\nu)$ for $\Gamma$ with marginals $\mu, \nu$, and a likelihood function that relates $\Gamma$ to a given sample $C^k$ of the cost. As we will mention below, priors on the transport polytope have already been discussed in the literature. Our key contribution is introducing the likelihood, quantifying how likely a given transport plan $\Gamma$ is optimal for a given cost matrix $C^k$.

\textbf{The Likelihood for $C^k$} is defined using \emph{auxiliary optimality variables} $O_k$ inspired by maximum entropy reinforcement learning~\citep{levine18}: define a binary variable $O_k\in \{0,1\}$ indicating whether $\Gamma$ achieves the minimum in $\OT(\mu,\nu, C^k)$ when $O_k=1$, and $O_k=0$ otherwise, for which we consider the distribution
\begin{equation}\label{eq:gamma_likelihood}
    \pr(O_k = 1 \mid \mu, \nu, C^k, \Gamma) = \exp\left(-\OT(C^k, \Gamma)\right),
\end{equation}
which allows writing the posterior in the form
\begin{equation}
\begin{aligned}
&\pr(\Gamma \mid \mu, \nu, O_k=1, C^k) \\
\propto \, & \pr(O_k=1 \mid \mu,\nu, C^k, \Gamma)\pr(\Gamma\mid \mu,\nu).
\end{aligned}
\end{equation}

This likelihood is motivated by the fact that $\OT(\mu,\nu, C^k)\geq 0$ always holds, and so if $\OT(C^k,\Gamma)=0$, then the likelihood of $\Gamma$ being optimal for $C^k$ (that is, $O_k = 1$) is $1$, as no lower value can be obtained. On the other hand, as $\OT(C^k, \Gamma)$ decreases, the likelihood increases. This effect is precisely what we wish for, as a lower total price indicates that $\Gamma$ is more optimal.

\textbf{Prior for $\Gamma$.} Any prior whose support covers the transport polytope could be used, such as the well-behaved ones discussed by \cite{frogner19}: component-wise normal, gamma, beta, chi-square, logistic and Weibull distributions. The authors also considered the Dirichlet distribution, which we find to work well in practice in the experimental section. We also consider the entropy prior, defined as 
\begin{equation}\label{eq:entropy_prior}
\pr(\Gamma)\propto \exp(\epsilon H(\Gamma)),\quad \epsilon > 0,
\end{equation}
which we use to enforce the positivity of the OT plans.
 
 \textbf{Posterior for $\Gamma$.}
 For a population of cost matrices $C^k$, $k=1,...,N$, we have two natural ways to define the posterior likelihood for $\Gamma$, assuming $O_k$ are independent and disjoint events. We either consider transport plans that are as optimal as possible \emph{for all} of the observed cost matrices, or, we require the transport plan to be optimal for \emph{some} of the cost matrices. These two choices lead to the following conditions:
 
 \begin{itemize}
     \item[(C1)] Condition $\Gamma$ on $O_k=1$ for \textbf{each} $k=1,..,N$.
     \item[(C2)] Condition $\Gamma$ on $O_k=1$ for \textbf{some} $k=1,...,N$.
 \end{itemize}
 As a short-hand notation, we denote the resulting posterior distributions, respectively, as 
\begin{equation}
\begin{aligned}
&\pr^\forall_{\mu,\nu}(\Gamma|C, O=1) \\
\overset{\Delta}{=} &\pr(\Gamma | \mu, \nu, C^k,~\forall k:~ O_k=1),\\
&\pr^\exists_{\mu,\nu}(\Gamma|C, O=1) \\
\overset{\Delta}{=} &\pr(\Gamma | \mu, \nu, C^k,~\exists k: ~O_k=1).
\end{aligned}
\end{equation}

 In practice, the events $O_k$ might not be independent, as arbtitrarily many $C_k$ might admit single $\Gamma$ as their minimizer. However, this assumption allows approximating the posterior likelihoods. For the condition (C1) we get the negative posterior log-likelihood
 \begin{equation}\label{eq:posterior_likelihood_forall}
\begin{aligned}
Q_\forall(\Gamma)=&-\log \pr^\forall_{\mu,\nu}(\Gamma\mid C, O=1)\\
=& - \log \left(\pr(\Gamma \mid \mu,\nu)\prod_{k=1}^N \pr(O_k = 1 \mid \mu, \nu, C^k, \Gamma)\right)\\
=& - \log \pr(\Gamma) + \OT\left(\sum_{k=1}^N C^k,\Gamma\right) + \mathrm{const.},
\end{aligned}
\end{equation}
and  for the condition (C2) we compute
\begin{equation}\label{eq:posterior_likelihood_exists}
\begin{aligned}
Q_\exists(\Gamma) =& -\log \pr^\exists_{\mu,\nu}(\Gamma|C, O=1)\\
=& - \log \left(\pr(\Gamma \mid \mu,\nu)\sum_{k=1}^N \pr(O_k = 1 \mid \mu, \nu, C^k, \Gamma)\right)\\
=& - \log \Pr(\Gamma) - \log \sum_{k=1}^N \exp(-\OT(C^k,\Gamma))\\
&+ \mathrm{const.}.
\end{aligned}
\end{equation}
The conditions lead  to quite different posterior likelihoods: both have the negative log prior-likelihood as a term, but the second terms differ. $Q_\forall$ has the average OT quantity over all the cost matrices, whereas $Q_\exists$ has a smooth minimum over the OT quantities.

\subsection{Posterior Sampling}\label{sec:posterior_sampling}
We consider a Markov chain Monte Carlo (MCMC), specifically a Hamiltonian Monte Carlo (HMC) method to sample from the OT plan posteriors. This requires a novel way to take the marginal constraints into account, which we do by utilizing the chart in~\eqref{eq:Gamma_chart}.

\textbf{MCMC} methods are the main workhorse behind Bayesian inference, allowing sampling from a given unnormalized distribution.

First,  a proposal process $\Pr(\Gamma_{t+1} | \Gamma_t)$ is devised. Given a proposed transition $\Gamma_t \to \Gamma_{t+1}$,  we filter it through the \emph{Metropolis-Hastings sampler}, ensuring that the resulting Markov chain is reversible with respect to $\Pr_{\mu,\nu}(\Gamma|C,O=1)$ and satisfies \emph{detailed balance}.

\textbf{HMC} is a celebrated variant of MCMC, allowing for efficient sampling in high dimensions, which pairs the state $\Gamma$ with a \emph{momentum} $P\in \R^{n\times m}$~\citep{neal11}. One then defines the \emph{kinetic energy} $T$ and \emph{potential energy} $U$, whose sum forms the \emph{Hamiltonian}
\begin{equation}
\Ham(\Gamma, P) = T(\Gamma,P) + U(\Gamma),
\end{equation}
which induces the \emph{Hamiltonian system}
whose trajectories preserve the Hamiltonian. The HMC procedure then samples a momentum $P_t$, and evolves the pair $(\Gamma_t, P_t)$ according to the Hamiltonian with a symplectic integrator, e.g., the \emph{leapfrog algorithm}. The resulting pair $(\Gamma_{t+1}, P_{t+1})$ is then accepted with probability
\begin{equation}\label{eq:HMC_acceptance}
\begin{aligned}
&\alpha(\Gamma_t, \Gamma_{t+1})\\
=& \min\left\lbrace 1, \exp\left(\Ham(\Gamma_t, P_t) - \Ham(\Gamma_{t+1}, P_{t+1})\right) \right\rbrace.
\end{aligned}
\end{equation}

\textbf{Constraints on $\Gamma$}, given in \eqref{eq:Gamma_constraints}, can be taken into account by parameterizing $\Gamma$ using the chart in \eqref{eq:Gamma_chart}. We also account for the positivity constraints $\Gamma\geq 0$ coordinate-wise by adding a small entropy term (with small $\epsilon$) defined in \eqref{eq:entropy_prior}, in the prior, so that any $\Gamma$ with negative values are rejected by the sampler in \eqref{eq:HMC_acceptance}, as the entropy would not be defined. That is, we propose writing $\pr(\Gamma) = \pr_1(\Gamma)\pr_2(\Gamma)$, where $\pr_1$ is the entropy prior, and $\pr_2$ an informative prior of our choosing. Alternatively, we could choose $\Pr_1(\Gamma)$ as the uniform distribution over the probability simplex.

\subsection{Maximum A Posteriori Estimation as Regularized OT}\label{sec:map_estimate}

We now consider the MAP estimate for the posterior distribution $\pr^\forall_{\mu,\nu}(\Gamma | C, O=1)$ under the condition (C1). The MAP estimate for condition (C2) is more demanding due to the non-convexity of the smooth minimum appearing in $Q_\exists$, whereas $Q_\forall$ is convex if $-\log \pr(\Gamma)$ is convex. Now considering $Q_\forall$ in \eqref{eq:posterior_likelihood_forall}, we see that computing the MAP estimate
\begin{equation}\label{eq:map_estimate}
\begin{aligned}
\Gamma_\forall^* =& \argmin\limits_{\Gamma \in \C(\mu,\nu)} Q_\forall(\Gamma)\\
=& \argmin\limits_{\Gamma \in \C(\mu,\nu)}\left\lbrace -\log \pr(\Gamma) + 
\OT\left( \sum_{k=1}^N C^k, \Gamma \right) \right\rbrace
\end{aligned}
\end{equation}
is equivalent to solving the regularized OT problem~\eqref{eq:rot_mover} with the regularizer $R(\Gamma) = -\log \pr(\Gamma)$ , the marginals $\mu,\nu$, and the cost matrix $\sum_k C^k$.

For the sake of illustration, we discuss the MAP estimate in three example cases.

\textbf{Constant Prior.} With a constant prior $\pr(\Gamma) = \mathrm{const.}$, solving \eqref{eq:map_estimate} corresponds to the vanilla OT problem~\eqref{eq:OT}.

\textbf{Entropy Prior.} Assume we have a prior proportional to the exponential of the $\epsilon$-scaled entropy of $\Gamma$ defined in~\eqref{eq:entropy_prior}, we get the regularizer
\begin{equation}
R(\Gamma) = -\log\pr(\Gamma) = -\epsilon H(\Gamma).
\end{equation}
Thus, solving \eqref{eq:map_estimate} corresponds precisely to solving the entropy-relaxed OT problem~\citep{cuturi13}.

\textbf{Gaussian Prior.}
Consider a Gaussian prior $\vect(\Gamma) \sim \pr(\Bar{\Gamma}, \Sigma)$ for the vectorized transport plan, with mean $\Bar{\Gamma}$ and covariance matrix $\Sigma$. Then, one gets
\begin{equation}
R(\Gamma) =  \frac{1}{2}(\vect(\Gamma)-\Bar{\Gamma}) \Sigma^{-1} (\vect(\Gamma) - \Bar{\Gamma}),
\end{equation}
and so if $\Bar{\Gamma}=0$, the Gaussian prior results in quadratically regularized OT~\citep{lorenz19,dessein18}, where the quadratic term is the norm with respect to the \emph{Mahalanobis metric} given by $\frac{1}{2}\Sigma^{-1}$.

\begin{table*}[h]
    \centering
    \small
    \begin{tabular}{lcccccccc}
        \toprule
         &\multicolumn{4}{c}{\textbf{No Cost}} & \multicolumn{4}{c}{\textbf{With Cost}} \\ 
         \cmidrule(lr){2-5} \cmidrule(lr){6-9}
         \textbf{Prior} & Error & Correlation & 1 STD & 2 STD& Error & Correlation & 1 STD & 2 STD  \\
         \midrule
         Dirichlet & $\mathbf{1.92\times 10^{-3}}$& $\mathbf{0.702}$& $\mathbf{62.8\%}$ & $\mathbf{82.9\%}$ & $1.91\times 10^{-3}$& $0.686$& $61.6\%$ & $82.4\%$ \\
         Tsallis & $2.62\times 10^{-3}$& $0.304$ & $47.9\%$ & $62.8\%$ & $2.54\times 10^{-3}$& $0.350$ & $47.1\%$ & $60.8\%$  \\
         Entropic & $2.44\times 10^{-3}$& $0.319$ & $45.4\%$ & $60.4\%$& $2.58\times 10^{-3}$& $0.240$ & $47.4\%$ & $63.0\%$  \\
         Gaussian & $2.41\times 10^{-3}$& $0.340$ & $44.3\%$ & $59.1\%$& $2.42\times 10^{-3}$& $0.322$ & $46.5\%$ & $60.7\%$  \\
         Uniform & $2.52\times 10^{-3}$ & $0.284$ & $44.3\%$ & $58.3\%$& $2.46\times 10^{-3}$ & $0.290$ & $48.2\%$ & $61.9\%$ \\
         \bottomrule
    \end{tabular}
    \caption{BayesOT yields meaningful uncertainty estimates for the Florida vote registration dataset. The \emph{median} error is computed for the mean posterior prediction, the correlation is between standard deviations of the posterior (for an entry in the joint distribution) and the absolute error, and the two last columns give the amount of data points lying inside the confidence bounds given by 1 and 2 standard deviations, respectively. The first four columns omit the OT likelihood term, whereas the four last columns include it.}
    \label{tab:ecological}
\end{table*}

\begin{figure}[H]
    \centering
    \includegraphics[width=\linewidth]{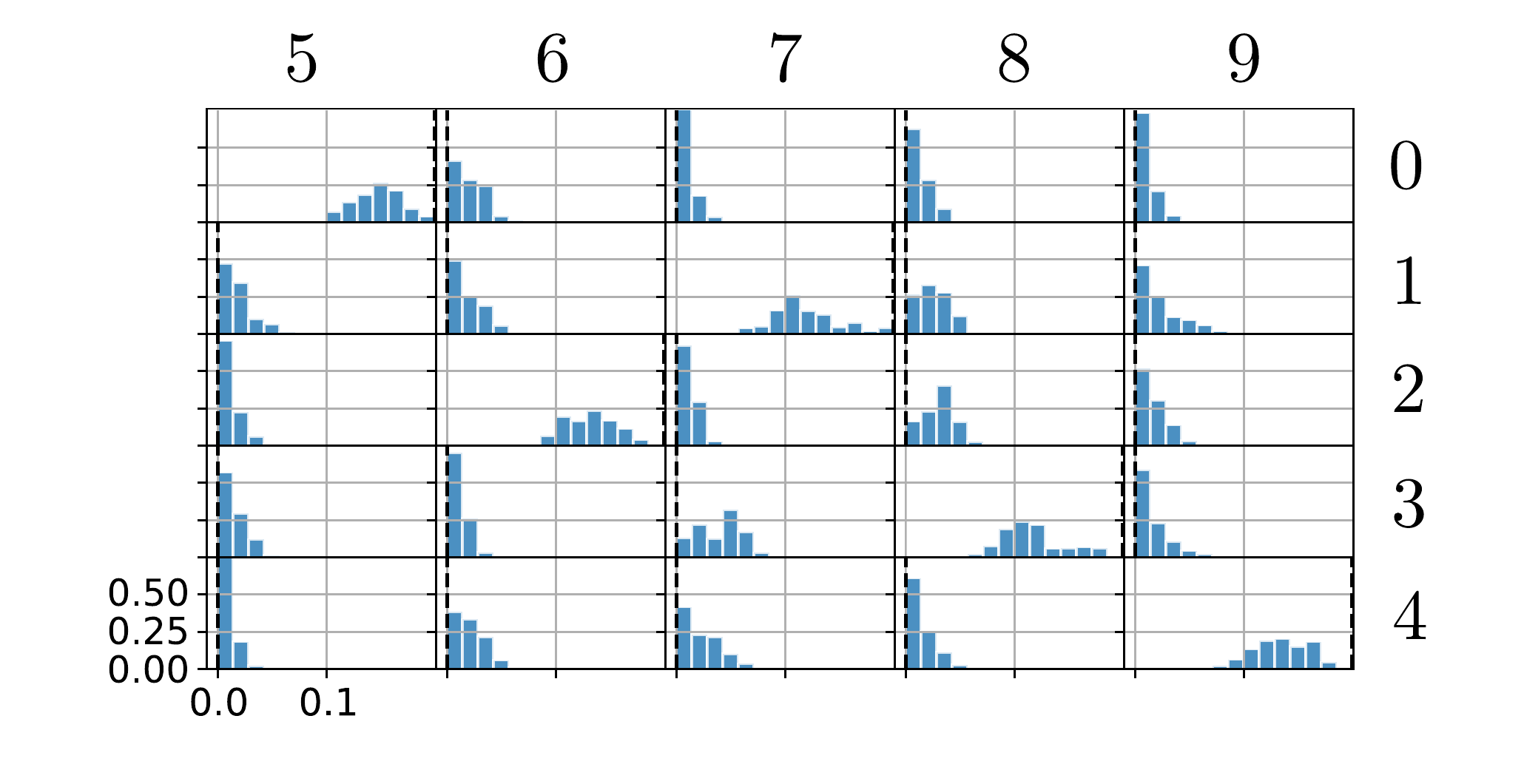}
    \caption{
    Demonstration of BayesOT between instances of digits 5-9 (columns) and 1-4 (rows). Each histogram shows the posterior of $\Gamma_{ij}$.}
    \label{fig:MNIST}
\end{figure}

\section{EXPERIMENTS}
We now demonstrate BayesOT on one toy data set (MNIST) and give empirical results on two sets: Florida vote registration dataset shows how BayesOT provides useful uncertainty estimates while building on top of traditional OT approaches. The New York City taxi dataset presents real traffic data, which we use to transport persons around Manhattan, comparing the BayesOT posterior to the average case analysis.

We implement BayesOT with the Pyro probabilistic programming framework~\citep{bingham19}, and use the NUTS sampler~\citep{hoffman14} for HMC to automatically tune the hyperparameters.

\textbf{MNIST.} As a toy-example with real data,  we consider transport between two measures over $32\times 32$ images of hand-written digits in the MNIST dataset~\citep{lecun98}. The digits $0-9$ are arbitrarily split into two groups of $0-4$ and $5-9$, forming two measures $\mu$ and $\nu$ with uniform weights. We sample images of each digit from the dataset to compute $N=100$ samples from the stochastic cost matrix using the squared Euclidean metric. We sample $10^4$ points from the posterior with $10^3$ burn in samples with a step size of $10^{-4}$, and use the entropy prior with $\epsilon=1$. 

The resulting posterior over the transport plans, conditioned on (C1) presented in Sec.~\ref{sec:ot_bayes_formulation}, is illustrated in Fig.~\ref{fig:MNIST}. The results positively match intuition, as we most often see the mappings $0\mapsto 5$, $1\mapsto 7$, $2\mapsto 6$, $3\mapsto 8$ and $4\mapsto 9$. However, some of the assignments are not as clear-cut as others. $0\mapsto 5$ is very dominant, whereas $3\mapsto 8$ is not that dominant, as in some cases $3\mapsto 7$ might be more favorable, depending on the drawing style of the digit.

\textbf{Florida Vote Registration.} We apply BayesOT to infer a joint table given two marginals, a common task in ecological inference. On top of point estimates, BayesOT provides uncertainty estimates, which are shown to be meaningful by the experiment.

The Florida dataset~\citep{imai16} describes $\approx 10^6$ individual voters in Florida for the 2012 US presidential elections. From the data, we aggregate two marginals per county (of which there are 68), namely a marginal of the party vote ('Democrat', 'Republican', 'Other') and another for ethnicity ('White', 'Black', 'Hispanic', 'Asian', 'Native', 'Other'). Then, we infer a posterior over joint tables between these features~\citep{flaxman15}, which we compare to ground truth joint tables for each county.

\cite{muzellec16} apply OT to this problem by using side information to compute a cost matrix as
\begin{equation}
C_{ij} = \sqrt{2 - 2\exp\left(-\gamma \|v_i^p - v_j^e\|_2\right)},
\end{equation}
where $\gamma = 10$, $v_i^p$ is the average profile for party $i$ of age normalized to lie within $[0,1]$, gender represented as a binary number and whether they voted in 2008 or not. $v_j^e$ is the same profile, but for ethnicity $j$.  \cite{muzellec16} employ \emph{Tsallis-regularized OT} to infer the joint table, which in our framework can be viewed as a MAP estimate with Tsallis-entropy prior. We show here how BayesOT, even when the cost is exact, allows us to provide uncertainty estimates for regularized OT, including Tsallis-regularized OT.

\begin{figure}
    \centering
    \includegraphics[width=1\linewidth]{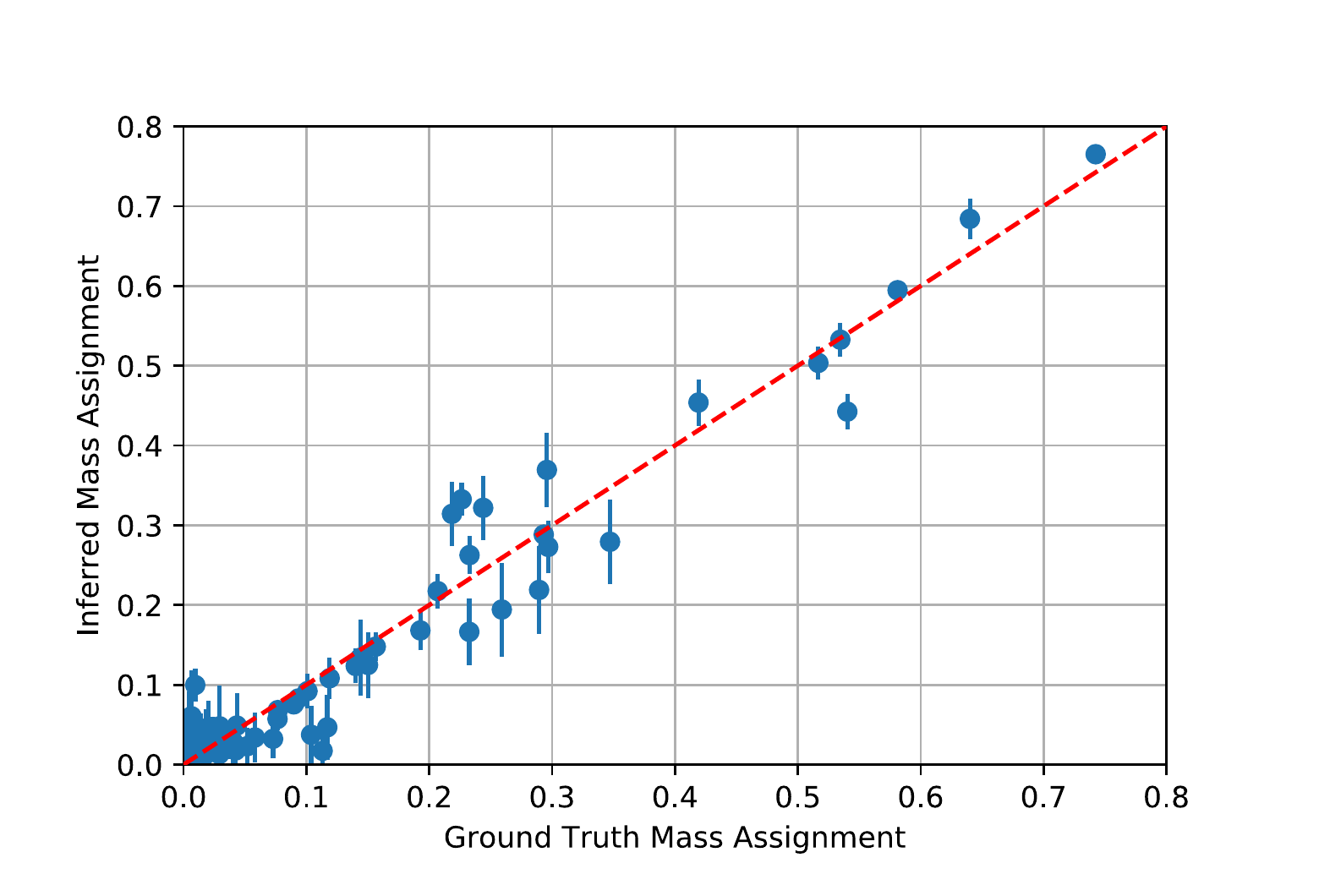}
    \caption{Ground-truth assignments against the posterior mean assignments $\bar{\Gamma}_{ij}$ for 10 first counties in the Florida vote registration dataset. The posterior utilizes Dirichlet prior with the cost matrix computed over the individual counties. A perfect inference would produce a scatter plot lying on the red diagonal line.}
    \label{fig:ecological}
\end{figure}

The approach by \cite{frogner19} discussed in Sec.~\ref{sec:intro} is also related. They choose a prior distribution, whose most likely joint table is chosen. Our HMC approach, which takes the marginal constraints into account, can then be applied to their work, by sampling from the prior distribution, yielding uncertainty estimates for the point estimate.

For each county, we vary the prior distribution between the Diriclet prior, the Tsallis-entropy prior and the entropy prior, and choose whether to use the likelihood associated with the OT cost or not (second term in \eqref{eq:posterior_likelihood_exists} and \eqref{eq:posterior_likelihood_forall}). In each case, the HMC chain is initialized with $10^2$ burn in samples with an initial step size of $10^{-4}$, after which $10^3$ posterior samples are acquired. This amount of samples is quite low, especially for higher dimensions, but the results show that meaningful uncertainty estimates are still obtained. 

The results are summarized in Table~\ref{tab:ecological}, presenting the median error, and to assess the uncertainty estimates, the correlation between uncertainty estimates and absolute error, and how many test values lie within the 1 STD and 2 STD confidence intervals of the point estimate. Furthermore, the results obtained using the Dirichlet prior and the cost matrix on the 10 first counties is illustrated in Fig.~\ref{fig:ecological}.

The results indicate clearly that the Dirichlet prior performs the best, as it achieves the lowest median error and highest correlation between the posterior standard deviations and absolute errors. This might be as the prior is supported on the probability simplex, and thus concentrates more mass there compared to the other priors. On the other hand, it is surprising that the cost matrix does not seem to provide meaningful information, as the results over each prior remain quite unaffected when we leave the OT likelihood term out.

\begin{figure}
    \centering
    \includegraphics[width=1\linewidth]{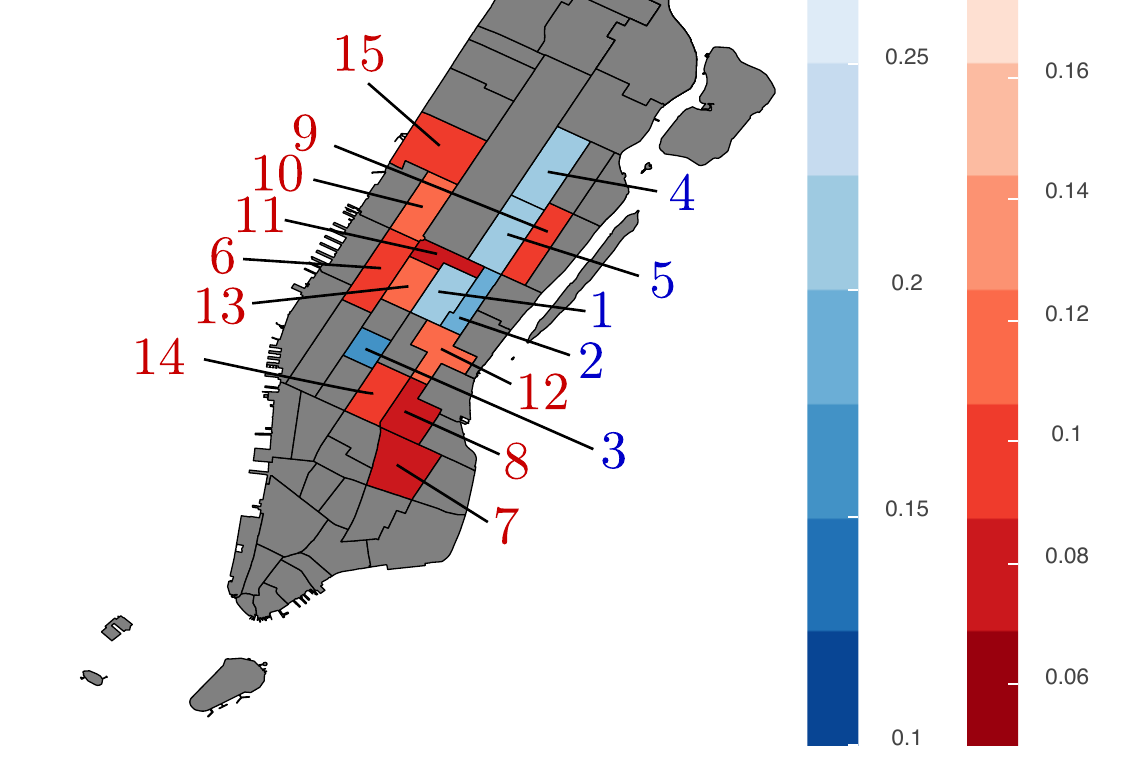}
    \caption{Mass distributions for used taxi zones on Manhattan. $\mu$ covers the blue zones, whereas $\nu$ covers the red zones.}
    \label{fig:manhattan_map}
\end{figure}

\textbf{NYC Taxi Dataset.} We consider data collected from Yellow cabs driving in Manhattan in January 2019, totalling 7.7 million trips. For $\mu$, we consider the 5 most common pick-up zones, and for $\nu$ the 6-15 most common pick-up zones, presented in Fig.~\ref{fig:manhattan_map}. The weights for $\mu$ (and $\nu$) are computed according to the amount of trips departing (and arriving) from the location. The cost matrix $C_{ij}$ is computed by sampling trips between locations $i$ and $j$, and dividing the fare by the amount of passengers on board. Thus, our task is to transport persons from pick-up locations to drop-off locations in an optimal way.

\begin{figure}
    \centering
    \includegraphics[width=1\linewidth]{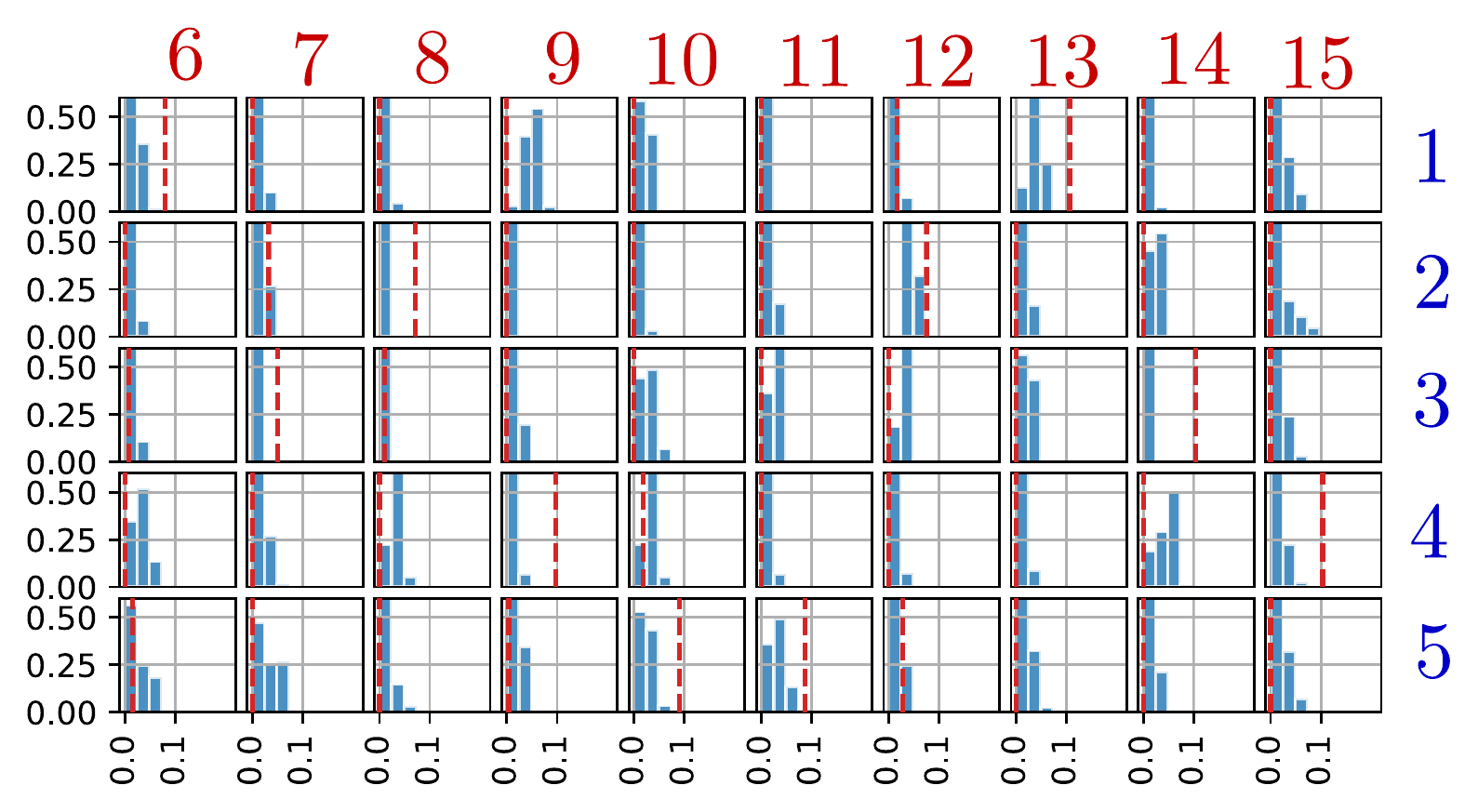}
    \caption{BayesOT posterior with uniform prior for transport plans between zones 1-5 and 6-15. Each histogram shows the posterior of $\Gamma_{ij}$, and the red lines give the standard OT solution for the average case.}
    \label{fig:taxi_posterior}
\end{figure}

For this experiment, we pick the uniform prior and obtain $1000$ samples from the stochastic cost matrix. We initialize the HMC chain with $2\times 10^3$ warm-up iterations, after which we sample $10^4$ points from the posterior, induced by (C2), which is illustrated in Fig.~\ref{fig:taxi_posterior}, alongside with the average cost OT solution.

In many cases where the average case analysis assigns considerable mass (e.g., $1\mapsto 13, 2\mapsto 12, 5\mapsto 11)$, we see a larger variation in the histogram towards larger mass assignments. This agrees with intuition, as there should be many individual cost matrices encouraging a large assignment, if the average OT plan has a large assignment. However, the histogram also supports low assignments, implying that it is not always optimal to match these taxi zones together. We do also observe contradicting cases, such as $3\mapsto 14$, which might be caused by a situation, where the assignment on average is optimal, but otherwise is not. On the other end, we also observe cases where on the average no mass is assigned ($1\mapsto 9, 3\mapsto 10$), but the histogram still tends to assign some mass. This could be caused by a similar case as above, where on average this is suboptimal, but in many cases one should still assign some mass.

\section{DISCUSSION}
We introduced BayesOT, an approach for studying OT with stochastic cost with Bayesian inference. The experiments endorse BayesOT as a successful approach to model the stochasticity that propagates to the OT plans from the cost, and even proves to be useful in providing uncertainty estimates for use cases of OT where an exact cost is used. 

A notable bottleneck for the use of BayesOT is formed by the posterior sampling method used. As we consider marginal distributions with an increasing amount of atoms, also the dimensionality of the problem increases, subsequently increasing the mixing time for the MCMC method used. Without notable improvements on the sampler, this prevents scaling BayesOT to large scale problems, although many use cases can be found in smaller problems, as we have demonstrated. A possible alternative to HMC could be the stochastic gradient Riemann Hamiltonian Monte Carlo~\citep{ma15}.

Possible future directions for BayesOT could include modelling the joint distribution $(C,\Gamma)$ of the cost and the OT plan explicitly, which allows computing a posterior distribution for the total OT cost. One could also consider regression problems, where at a given time with no observations, a distribution over potential OT plans could be inferred based on previous data. Although advances are needed, based on the experiments, we view BayesOT as a useful first step towards making OT-based analysis possible in uncertain environments.

\subsubsection*{Acknowledgements}
This work was supported by the Academy of Finland (Flagship programme: Finnish Center for Artificial Intelligence FCAI, Grants 294238, 319264, 292334, 334600, 324800). We acknowledge the computational resources provided by Aalto Science-IT project.


{\small
\bibliographystyle{apalike}
\bibliography{references}
}
\end{document}